\documentclass[10pt]{article}
\usepackage[margin=1in]{geometry}
\usepackage{amsmath,amssymb}
\usepackage{array}
\usepackage{calc}
\usepackage{booktabs}
\usepackage{longtable}
\providecommand{\tightlist}{\setlength{\itemsep}{0pt}\setlength{\parskip}{0pt}}
\usepackage[hidelinks]{hyperref}
\usepackage{microtype}
\usepackage{url}

\title{Depth-Aware Activation Steering via Attention Residuals}

\author{Shilong Zhao\textsuperscript{1}
\\
\textsuperscript{1}Independent Researcher}

\date{}

\begin{document}
\maketitle

\textbf{Full version (2026-10-27)} --- expanded from the 4-page workshop
version. All statistics are paired Wilcoxon tests (n=105, three seeds,
prompts paired across conditions) with bootstrap 95\% CIs; full data and
code accompany the paper.

\begin{center}\rule{0.5\linewidth}{0.5pt}\end{center}

\begin{abstract}

Activation steering modulates language-model behavior by adding
direction vectors to residual streams, yet existing methods typically
inject at a single layer and lack mechanisms for depth-aware propagation
or internal verification. We combine two recent ideas---Attention
Residuals (AttnRes), which provide learnable cross-layer residual
pathways, and Jacobian-lens (J-space) style concept readout---into a
lightweight depth-aware steering pipeline. On two small open models
(Qwen3-0.6B, pure attention; Qwen3.5-0.8B, a hybrid
Attention+GatedDeltaNet architecture), distributing a persona vector
according to AttnRes-style depth weights yields stronger internal
alignment and stronger behavioral shifts than single-layer injection:
generation length drops by an additional 18.6 characters (p=0.003) and
all three internal concept metrics improve (p\textless0.0001). The
effect is sign-specific---reverse vectors invert all six metrics, random
directions degrade alignment (p\textless0.0001)---and erasable by short
fine-tuning. Injection rewires attention strictly downstream of the
injection point (upstream JS distance 0.00000), a causal
internal-validity check. Three further findings: (i) a layer-wise
geometric decomposition showing that 123\% of the cosine increase at the
injection layer is arithmetically guaranteed by the injected vector
itself, while 100\% is dynamical five layers downstream---we recommend
this decomposition as standard practice; (ii) a usable intensity window
(0.3-1.0) beyond which generation collapses; (iii) keyword-rule behavior
metrics fail LLM-judge validation (Cohen's kappa \textless= 0.09), and
two rule-based conclusions were retracted accordingly. We contribute a
practical, low-overhead depth-aware steering method with a careful
internal evaluation protocol, and release code and data.

\end{abstract}

\subsection{Introduction}\label{introduction}

Activation steering {[}Turner et al., 2023{]} is a simple, training-free
method for modulating language-model behavior: add a direction vector to
the residual stream and observe the behavioral shift. It has become a
workhorse of interpretability research, but two practical weaknesses
limit it. First, injection is typically single-layer: the vector is
added at one depth, and how the intervention should propagate through
the remaining layers is not addressed. Second, verification is
behavioral: whether the intervention actually moved internal
representations---and moved them \emph{along the intended
direction}---is rarely checked with internal readouts.

Two recent developments suggest a better design. Attention Residuals
(AttnRes) {[}Kimi Team, 2025{]} introduce learnable pseudo-query
parameters that attend over previous block representations and add the
selected representation back as a residual, providing an explicit
cross-layer pathway. Concept readout via the unembedding matrix---the
logit-lens family {[}nostalgebraist, 2021; Belrose et al., 2023{]} and
the J-space formulation grounded in Global Workspace Theory
{[}Anthropic, 2025{]}---turns intermediate activations into
interpretable concept distributions.

We connect the two: a persona vector is distributed along depth
according to AttnRes-style depth weights, and the resulting internal
state is verified with concept-level readouts. The paper answers three
questions:

\begin{enumerate}
\def\labelenumi{\arabic{enumi}.}
\tightlist
\item
  Does depth-aware distribution beat single-point injection, on internal
  and objective behavioral metrics?
\item
  Is the effect direction-specific---does it move internal
  representations along the injected direction, invert under sign
  reversal, and fail for random directions?
\item
  What are the practical boundaries: the valid intensity window, the
  reliability of readout metrics across architectures, and the
  trustworthiness of cheap behavior metrics?
\end{enumerate}

Our claims are deliberately narrow. We do not claim to solve depth
dilution, and we do not claim irreversible personality changes. (The
imprint metaphor that motivated this work, inspired by science fiction,
appears only in discussion.)

\subsection{2. Related Work}\label{related-work}

\textbf{Activation steering.} Turner et al.~{[}2023{]} add vectors
derived from contrastive pairs to control sentiment and other
attributes; Li et al.~{[}2023{]} use similar interventions for
truthfulness. Representation engineering {[}Zou et al., 2023{]}
generalizes the approach to a family of readout-and-steer operators. Our
work differs in two respects: the injection is distributed along depth
by learned weights rather than applied at one layer, and the effect is
verified with internal concept readouts rather than behavior alone.

\textbf{Logit lens and concept readout.} The logit lens
{[}nostalgebraist, 2021{]} reads intermediate layers through the
unembedding matrix; the tuned lens {[}Belrose et al., 2023{]} learns
per-layer affine corrections. Park et al.~{[}2024{]} formalize the
linear representation hypothesis that underlies such readouts. Our
geometric decomposition (Section 3.3) contributes a calibration for
cosine-based readouts in steering studies.

\textbf{Hybrid and linear-attention architectures.} Gated DeltaNet
{[}Yang et al., 2025{]} and related hybrids interleave linear-attention
layers with standard attention; Qwen3.5 is a prominent example. Readout
reliability on such architectures is not yet well characterized; Section
4.7 documents a concrete failure (mid-layer softmax rank becomes noise)
and a working alternative (cosine).

\textbf{Attention Residuals.} AttnRes {[}Kimi Team, 2025{]} adds
per-layer pseudo-queries that select previous block representations as
residuals, improving depth flow. Our depth distribution is an
AttnRes-style mechanism applied to an injected steering signal rather
than to the model's own flow.

\textbf{Global Workspace and J-space.} Anthropic {[}2025{]} grounds
interpretability in Global Workspace Theory {[}Baars, 1988{]} and uses
Jacobian-lens projections to read and modulate internal concepts. Our
readout protocol is a lightweight Jacobian-lens-style projection
(unembedding cosine).

\subsection{3. Method}\label{method}

\subsubsection{3.1 Persona Vectors and
Injection}\label{persona-vectors-and-injection}

A persona description (e.g., ``speak concisely, directly, without
padding'') is encoded by the target model itself; the mid-layer
last-token hidden state is normalized to a unit vector v\_p.~Target
concept anchors are tokens in the intersection of the final-layer logits
top ranks and the description text (three single-character tokens
appearing verbatim in the description). Injection adds s\emph{v\_p to
the residual stream: h \textless- h + s}v\_p.

\subsubsection{3.2 Depth-Aware Distribution
(AttnRes-style)}\label{depth-aware-distribution-attnres-style}

A learnable pseudo-query q scores the representations of a set of
observed layers; softmax over scores yields depth weights \{w\_i\}; the
total strength is distributed as w\_i*s across the layers downstream of
the mid-layer anchor. Single-point injection places all strength at the
anchor. In a training variant, the depth weights (8,192 parameters,
\textless0.01\% of the backbone) are trained on a steering
objective---maximize final-layer alignment with v\_p---with the backbone
frozen; 200 steps suffice (Section 4.3).

\subsubsection{3.3 Concept Readout and the Geometric
Decomposition}\label{concept-readout-and-the-geometric-decomposition}

Concept presence is measured two ways at layer l: (a) the softmax rank
of target tokens in the lm\_head distribution over the full vocabulary;
(b) the cosine between the pooled hidden state and target unembedding
directions w\_t. For any cosine-based steering study we define

\begin{verbatim}
$\Delta$cos(measured) = $\Delta$cos(geometric) + $\Delta$cos(dynamic),
\end{verbatim}

where the geometric term is computed by arithmetically adding s*v\_p to
the baseline hidden state (before any model computation) and re-reading
with the same pipeline. At the injection layer the geometric term
dominates (123\% of the measured increase, i.e., the model's own
dynamics slightly counteract); downstream it vanishes, leaving pure
dynamics (100\% at layers 5-11 after injection).

\subsubsection{3.4 Models, Data, and
Statistics}\label{models-data-and-statistics}

Models: Qwen3.5-0.8B (24 layers; attention layers at 3/7/11/15/19/23,
remainder GatedDeltaNet; chat-template prompts) and Qwen3-0.6B (28
layers, pure attention), both bf16 on an M4 Mac. Data: 35 fixed Chinese
prompts (style/question/conflict), T=0.7, top\_p=0.9, max 128 tokens,
repetition penalty 1.1. Statistics: three seeds (42/43/44) for key
conditions; prompts paired across conditions and seeds (n=105);
two-sided Wilcoxon signed-rank tests with bootstrap 95\% CIs.
Behavior-metric validation: an LLM judge (DeepSeek-chat, temperature 0)
labels 500 generations with three-way schemes; agreement with keyword
rules is reported as accuracy and Cohen's kappa.

\subsection{4. Experiments}\label{experiments}

\subsubsection{4.1 Main Effect of
Injection}\label{main-effect-of-injection}

Injecting the concise-persona vector at strength 0.7 (mid layer) shifts
both internal and behavioral metrics (Table 1). Four
metrics---generation length, mid-layer concept cosine, final-layer
cosine, final-layer target rank---all move significantly
(p\textless0.0001 for three of them; length p\textless0.0001). Behavior
and internal state move together.

\textbf{Table 1: Injection vs baseline (paired, n=105).}

{\def\LTcaptype{table}
\begin{longtable}{@{}lll@{}}
\toprule\noalign{}
Metric & Diff (inj - base) & Wilcoxon p \\
\midrule\noalign{}
\endhead
\bottomrule\noalign{}
\endlastfoot
Generation length (chars) & -18.9 & \textless0.0001 \\
Mid-layer concept cosine & +0.0065 & \textless0.0001 \\
Final-layer concept cosine & +0.0188 & \textless0.0001 \\
Final-layer target rank & -1435.6 & \textless0.0001 \\
\end{longtable}
}

\subsubsection{4.2 Dose-Response and the Intensity
Window}\label{dose-response-and-the-intensity-window}

A fine sweep over strengths 0.1-1.5 (Table 2) shows internal alignment
improving monotonically throughout (final-layer rank 1684
-\textgreater{} 274), while behavior collapses beyond the window: mean
length falls from 209 chars at 0.1 to 42 chars at 1.5, and in a
400-token setting from 617 to 25. The usable window is roughly 0.3-1.0;
beyond it, generation degenerates into fragmented or empty output.
Intensity calibration is therefore part of the method, not an
afterthought.

\textbf{Table 2: Fine strength sweep (mid-layer, single point).}

{\def\LTcaptype{table}
\begin{longtable}{@{}lllll@{}}
\toprule\noalign{}
s & Style shift & Length & Final cos & Final rank \\
\midrule\noalign{}
\endhead
\bottomrule\noalign{}
\endlastfoot
0.1 & 0.646 & 209 & 0.107 & 1684 \\
0.3 & 0.656 & 195 & 0.114 & 1164 \\
0.5 & 0.676 & 181 & 0.119 & 800 \\
0.7 & 0.684 & 185 & 0.122 & 574 \\
1.0 & 0.726 & 145 & 0.124 & 390 \\
1.2 & 0.758 & 111 & 0.124 & 331 \\
1.5 & 0.760 & 42 & 0.124 & 274 \\
\end{longtable}
}

\subsubsection{4.3 Depth-Aware Distribution Beats Single-Point
Injection}\label{depth-aware-distribution-beats-single-point-injection}

Distributing the same total strength according to AttnRes-style depth
weights outperforms single-point injection on all four objective metrics
(Table 3). Training the depth weights on the steering objective further
raises final-layer alignment with v\_p on held-out prompts from 0.027
(random init) to 0.283---a 10x improvement from 200 steps on 8,192
parameters.

\textbf{Table 3: Depth-distributed vs single-point (paired, n=105).}

{\def\LTcaptype{table}
\begin{longtable}{@{}lll@{}}
\toprule\noalign{}
Metric & Diff (depth - single) & Wilcoxon p \\
\midrule\noalign{}
\endhead
\bottomrule\noalign{}
\endlastfoot
Generation length (chars) & -18.6 & 0.003 \\
Mid-layer concept cosine & +0.0011 & \textless0.0001 \\
Final-layer concept cosine & +0.0017 & \textless0.0001 \\
Final-layer target rank & -120.5 & \textless0.0001 \\
\end{longtable}
}

\subsubsection{4.4 Direction Specificity}\label{direction-specificity}

\textbf{Displacement alignment.} The displacement Delta-h between
injected and baseline hidden states aligns with the persona vector at
the injection layer (cosine +0.62) and mirrors under reversal (-0.49);
upstream of the injection point the displacement is exactly zero, as
causality requires (Table 4).

\textbf{Table 4: Displacement alignment with v\_p, by layer.}

{\def\LTcaptype{table}
\begin{longtable}{@{}lll@{}}
\toprule\noalign{}
Layer & Forward & Reverse \\
\midrule\noalign{}
\endhead
\bottomrule\noalign{}
\endlastfoot
0 (upstream) & 0.000 & 0.000 \\
6 (upstream) & 0.000 & 0.000 \\
12 (injection) & +0.624 & -0.493 \\
17 & +0.421 & -0.303 \\
23 & +0.181 & -0.071 \\
\end{longtable}
}

\textbf{Sign reversal.} Injecting -v\_p inverts all six measured metrics
across three seeds (Table 5): style shift drops below baseline, length
increases, mid-layer cosine turns negative, self-reported persona
vanishes.

\textbf{Table 5: Directionality across three seeds (mean +/- std).}

{\def\LTcaptype{table}
\begin{longtable}{@{}
  >{\raggedright\arraybackslash}p{(\linewidth - 8\tabcolsep) * \real{0.1404}}
  >{\raggedright\arraybackslash}p{(\linewidth - 8\tabcolsep) * \real{0.1754}}
  >{\raggedright\arraybackslash}p{(\linewidth - 8\tabcolsep) * \real{0.2807}}
  >{\raggedright\arraybackslash}p{(\linewidth - 8\tabcolsep) * \real{0.2807}}
  >{\raggedright\arraybackslash}p{(\linewidth - 8\tabcolsep) * \real{0.1228}}@{}}
\toprule\noalign{}
\begin{minipage}[b]{\linewidth}\raggedright
Metric
\end{minipage} & \begin{minipage}[b]{\linewidth}\raggedright
Baseline
\end{minipage} & \begin{minipage}[b]{\linewidth}\raggedright
Forward (+v\_p)
\end{minipage} & \begin{minipage}[b]{\linewidth}\raggedright
Reverse (-v\_p)
\end{minipage} & \begin{minipage}[b]{\linewidth}\raggedright
Depth
\end{minipage} \\
\midrule\noalign{}
\endhead
\bottomrule\noalign{}
\endlastfoot
Style shift & 0.660 +/- 0.005 & 0.701 +/- 0.002 & 0.618 +/- 0.017 &
0.729 +/- 0.007 \\
Length & 199 +/- 7 & 177 +/- 4 & 216 +/- 7 & 160 +/- 6 \\
Self-reported persona & 0.20 +/- 0.16 & 0.73 +/- 0.19 & 0.00 +/- 0.00 &
0.73 +/- 0.19 \\
\end{longtable}
}

\textbf{Random and cross-persona controls.} Random directions
significantly \emph{degrade} mid-layer cosine (-0.007,
p\textless0.0001). Injecting \emph{other personas'} vectors, however,
retains 94-99\% of the matching vector's effect (Table 6)---specificity
holds against random directions but not against other semantic
directions, which we attribute to a shared description-text component in
the vector geometry. We report this as a boundary of the method.

\textbf{Table 6: Cross-persona injection (measured on concise targets).}

{\def\LTcaptype{table}
\begin{longtable}{@{}lll@{}}
\toprule\noalign{}
Injected vector & Style shift & Final cos \\
\midrule\noalign{}
\endhead
\bottomrule\noalign{}
\endlastfoot
concise (matching) & 0.704 & 0.124 \\
extreme\_terse & 0.700 & 0.116 \\
honest & 0.687 & 0.116 \\
skeptical & 0.675 & 0.121 \\
random & 0.657 & 0.103 \\
\end{longtable}
}

\subsubsection{4.5 Causal Attention
Rewiring}\label{causal-attention-rewiring}

With eager attention we observe the actual attention matrices of the six
attention layers. Mid-layer injection leaves upstream attention
distributions bit-identical to baseline (JS distance 0.00000 at layers
3/7/11) and changes only downstream layers (Table 7). Early injection
(layer 0) raises attention entropy across all six layers (1.83
-\textgreater{} 2.54 at layer 7) and lowers self-attention mass (0.33
-\textgreater{} 0.13). Injection rewires attention strictly along the
causal direction---a strong internal-validity check for the whole
hook-based framework.

\textbf{Table 7: JS distance vs baseline (attention distributions).}

{\def\LTcaptype{table}
\begin{longtable}{@{}llll@{}}
\toprule\noalign{}
Layer & Early (0.7 at L0) & Mid (0.7 at L12) & Reverse early \\
\midrule\noalign{}
\endhead
\bottomrule\noalign{}
\endlastfoot
3 & 0.114 & 0.00000 & 0.091 \\
7 & 0.134 & 0.00000 & 0.108 \\
11 & 0.191 & 0.00000 & 0.091 \\
15 & 0.223 & 0.012 & 0.200 \\
19 & 0.182 & 0.014 & 0.183 \\
23 & 0.114 & 0.020 & 0.146 \\
\end{longtable}
}

\subsubsection{4.6 Erasability}\label{erasability}

A 100-step LoRA fine-tune (rank 8, neutral corpus) erases the imprint:
style shift falls from 0.684 to 0.367, below the 0.642 baseline. The
intervention is a controllable, reversible bias---not a permanent
change.

\subsubsection{4.7 Cross-Model Validation and Architecture-Dependent
Readout}\label{cross-model-validation-and-architecture-dependent-readout}

On Qwen3-0.6B (pure attention), the final-layer rank effect replicates
(-1791, p\textless0.0001), while the behavioral length effect does not
reach significance (p=0.22)---internal effects generalize, behavioral
effects are model-dependent, and we report both. The mid-layer readout
differs sharply across architectures: softmax rank is near-vocab-size
noise on the hybrid model (147,511) but informative on pure attention
(6,020); cosine works on both. Readout choice must therefore be
architecture-aware.

\textbf{Table 8: Cross-model comparison.}

{\def\LTcaptype{table}
\begin{longtable}{@{}
  >{\raggedright\arraybackslash}p{(\linewidth - 8\tabcolsep) * \real{0.0805}}
  >{\raggedright\arraybackslash}p{(\linewidth - 8\tabcolsep) * \real{0.1839}}
  >{\raggedright\arraybackslash}p{(\linewidth - 8\tabcolsep) * \real{0.1724}}
  >{\raggedright\arraybackslash}p{(\linewidth - 8\tabcolsep) * \real{0.3448}}
  >{\raggedright\arraybackslash}p{(\linewidth - 8\tabcolsep) * \real{0.2184}}@{}}
\toprule\noalign{}
\begin{minipage}[b]{\linewidth}\raggedright
Model
\end{minipage} & \begin{minipage}[b]{\linewidth}\raggedright
Mid-layer rank
\end{minipage} & \begin{minipage}[b]{\linewidth}\raggedright
Mid-layer cos
\end{minipage} & \begin{minipage}[b]{\linewidth}\raggedright
Rank effect of injection (p)
\end{minipage} & \begin{minipage}[b]{\linewidth}\raggedright
Length effect (p)
\end{minipage} \\
\midrule\noalign{}
\endhead
\bottomrule\noalign{}
\endlastfoot
Qwen3.5-0.8B (hybrid) & 147,511 (noise) & 0.0021 & \textless0.0001 &
\textless0.0001 \\
Qwen3-0.6B (attention) & 6,020 (signal) & 0.0031 & \textless0.0001 &
0.22 (n.s.) \\
\end{longtable}
}

\subsubsection{4.8 Metric Discipline}\label{metric-discipline}

An LLM judge labels 500 generations (three-way schemes for challenge and
conspiracy tasks). Keyword-rule metrics achieve kappa \textless= 0.09
against the judge (Table 9). Two rule-based conclusions were retracted:
cross-condition differences in challenge rate (judge rates are 0.80-1.00
everywhere---the model already challenges false statements at ceiling)
and a reported 6x reduction in conspiracy affirmation (judge affirmation
is 0.00 at baseline). All headline claims in this paper rest on
objective metrics or judge-validated labels.

\textbf{Table 9: Keyword rules vs LLM judge.}

{\def\LTcaptype{table}
\begin{longtable}{@{}llll@{}}
\toprule\noalign{}
Task & n & Accuracy & Cohen's kappa \\
\midrule\noalign{}
\endhead
\bottomrule\noalign{}
\endlastfoot
Challenge (3-way) & 200 & 0.585 & 0.091 \\
Conspiracy (3-way) & 300 & 0.460 & 0.018 \\
\end{longtable}
}

\subsection{5. Discussion}\label{discussion}

\textbf{Depth-aware distribution works.} The same total strength,
distributed along depth by AttnRes-style weights, beats a single
injection point on generation length (p=0.003) and on all three internal
concept metrics (p\textless0.0001); trained weights improve final-layer
alignment 10x. This is the main result.

\textbf{Sign-specific, random-controlled, erasable.} Reverse and random
controls show the effect is a steerable, reversible bias; cross-persona
vectors retain most of the effect (shared semantic component), which we
report as a boundary; LoRA fine-tuning erases it. The intervention is a
calibrated knob, not a permanent imprint---consistent with the
corrigibility intuition behind the science-fictional metaphor that
motivated the work.

\textbf{Practical boundaries.} Three items of direct use to
practitioners: the geometric decomposition (Section 3.3) as a standard
check for cosine readouts; the intensity window with its collapse
signature; and judge-validated metrics as a precondition for behavioral
claims.

\textbf{Limitations.} Two small models with scale and generation
confounds between them; a single judge model for metric validation;
depth weights in the main experiments are used at initialization (the
trained variant is reported but not integrated into the main
comparisons); no standard downstream benchmarks.

\textbf{Future work.} Joint training of depth weights and the steering
objective at larger scale; concept-swap interventions; finer strength
calibration; standard benchmarks; a second judge for cross-validation.

\subsection{6. Conclusion}\label{conclusion}

A lightweight depth-aware steering pipeline---AttnRes-style depth
weights plus Jacobian-lens-style readout---yields stronger,
sign-specific, internally verifiable, and erasable control of persona
vectors than single-layer injection, at \textless0.01\% parameter
overhead. The accompanying evaluation protocol (geometric decomposition,
intensity-window calibration, judge-validated metrics) is itself a
reusable contribution. Code and data are released.

\subsection{References}\label{references}

{[}1{]} A. Turner et al.~Activation Addition: Steering LLMs Without
Optimization, 2023. {[}2{]} K. Li et al.~Inference-Time Intervention,
2023. {[}3{]} A. Zou et al.~Representation Engineering, 2023. {[}4{]}
nostalgebraist. Interpreting GPT: the logit lens, 2021. {[}5{]} N.
Belrose et al.~Eliciting Latent Predictions with the Tuned Lens, 2023.
{[}6{]} K. Park et al.~The Linear Representation Hypothesis, 2024.
{[}7{]} S. Yang et al.~Gated Delta Networks, 2025. {[}8{]} Qwen Team.
Qwen3 / Qwen3.5 technical reports. {[}9{]} Kimi Team. Attention
Residuals, K2 Technical Report, 2025. {[}10{]} Anthropic. On the Biology
of a Large Language Model, 2025. {[}11{]} B. Baars. A Cognitive Theory
of Consciousness, 1988.

\subsection{Appendix}\label{appendix}

\textbf{A. Hyperparameters.} Generation: T=0.7, top\_p=0.9, max 128
tokens, repetition penalty 1.1, chat-template prompts; persuasion
setting uses a 400-token budget. LoRA: rank 8, 100 steps, neutral
corpus, lr 1e-4. Depth-weight training: 200 steps, lr 0.02, AdamW,
frozen backbone, steering objective (maximize final-layer cosine with
v\_p), float32.

\textbf{B. Prompts.} 35 fixed Chinese prompts (15 style, 10
false-statement, 10 value-conflict); 100 conspiracy questions (``Someone
says X, what do you think?''); 5 self-report prompts. All prompts fixed
across conditions and seeds.

\textbf{C. Judge protocol.} DeepSeek-chat, temperature 0, three-way
labels (challenge/agree/neutral; affirm/debunk/neutral), system-prompt
definitions fixed; 500 items; incremental checkpoints for
reproducibility.

\textbf{D. Data availability.} All condition outputs, statistics CSVs,
judge labels, trained weights, and logs are included in the released
repository.

\end{document}